\documentclass[letterpaper]{article} 
\usepackage{aaai23}  
\usepackage{times}  
\usepackage{helvet}  
\usepackage{courier}  
\usepackage[hyphens]{url}  
\usepackage{graphicx} 
\usepackage{natbib}  
\usepackage{caption} 
\usepackage{algorithm}
\usepackage{algorithmic}
\usepackage{graphicx}
\usepackage{amsmath}
\usepackage{amsthm}
\usepackage{booktabs}
\usepackage{amssymb}
\usepackage{multirow}
\usepackage{array}
\usepackage{setspace}
\usepackage{pifont}
\usepackage{adjustbox}
\usepackage{makecell}
\usepackage{color}
\usepackage{subfigure}
\usepackage{graphics}

\usepackage{newfloat}
\usepackage{listings}
\DeclareCaptionStyle{ruled}{labelfont=normalfont,labelsep=colon,strut=off} 
\floatstyle{ruled}
\newfloat{listing}{tb}{lst}{}
\floatname{listing}{Listing}
\title{Distantly-Supervised Named Entity Recognition with Adaptive \\  Teacher Learning and Fine-grained Student Ensemble}
\author{
    Xiaoye Qu,\textsuperscript{\rm 1}
    Jun Zeng\textsuperscript{\rm 2}, Daizong Liu\textsuperscript{\rm 3}, Zhefeng Wang\textsuperscript{\rm 1}\thanks{Zhefeng Wang and Pan Zhou are corresponding authors. The first two authors contribute equally.}, Baoxing Huai\textsuperscript{\rm 1}, Pan Zhou\textsuperscript{\rm 4}\textsuperscript{\rm *}
}

\affiliations{
    \textsuperscript{\rm 1}Huawei Cloud\\
    \textsuperscript{\rm 2}School of Software Engineering, Huazhong University of Science and Technology\\
    \textsuperscript{\rm 3}Peking University\\
    \textsuperscript{\rm 4}Hubei Key Laboratory of Distributed System Security, Hubei Engineering Research Center on Big Data Security, School of Cyber Science and Engineering, Huazhong University of Science and Technology 

    \{quxiaoye, wangzhefeng, huaibaoxing\}@huawei.com, dzliu@stu.pku.edu.cn,  
    \{\text{junzeng}, panzhou\}@hust.edu.cn
}

\usepackage{bibentry}

\begin{document}

\maketitle

\begin{abstract}
Distantly-Supervised Named Entity Recognition (DS-NER) effectively alleviates the data scarcity problem in NER by automatically generating training samples. Unfortunately, the distant supervision may induce noisy labels, thus undermining the robustness of the learned models and restricting the practical application. 
To relieve this problem, recent works adopt self-training teacher-student frameworks to gradually refine the training labels and improve the generalization ability of NER models. However, we argue that the performance of the current self-training frameworks for DS-NER is severely underestimated by their plain designs, including both inadequate student learning and coarse-grained teacher updating. Therefore, in this paper, we make the first attempt to alleviate these issues by proposing:  
(1) adaptive teacher learning comprised of joint training of two teacher-student networks and considering both consistent and inconsistent predictions between two teachers, thus promoting comprehensive student learning. 
(2) fine-grained student ensemble that updates each fragment of the teacher model with a temporal moving average of the corresponding fragment of the student, which enhances consistent predictions on each model fragment against noise. 
To verify the effectiveness of our proposed method, we conduct experiments on four DS-NER datasets. The experimental results demonstrate that our method significantly surpasses previous SOTA methods. The code is available at \text{https://github.com/zenhjunpro/ATSEN}.
\end{abstract}

\section{Introduction}

Named Entity Recognition (NER) aims to detect entity mentions in the text and classify them into predefined types, such as person, location, and organization. It is a fundamental task in information extraction and benefits many downstream NLP applications (e.g., relation extraction \cite{cheng2021hacred}, co-reference resolution \cite{clark-manning-2016-improving}, entity linking \cite{gu2021read} and event extraction \cite{ijcai2022p632}). In recent years, deep supervised models \cite{li2022unified,gu2022delving,li2020unified} have achieved superior success in the NER field. However, these supervised NER methods demand a large amount of high-quality annotation, which is extremely labor-intensive and time-consuming as NER demands token-level annotation.

\begin{figure}[t]
\centering
\includegraphics[width=7.8cm]{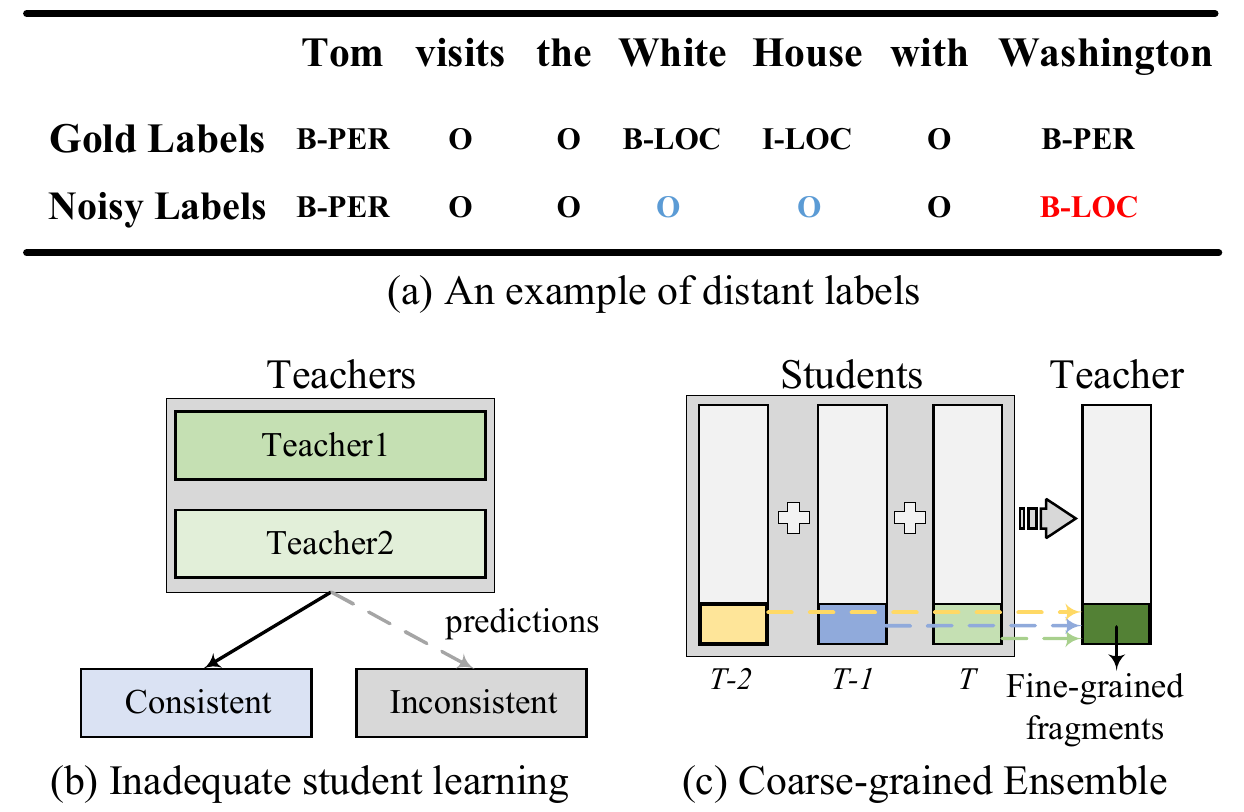}
\caption{(a) {\it White House} and {\it Washington} are incomplete and inaccurate labels. 
(b) Previous method only considers the consistent prediction parts from teachers, leading to incomprehensible student learning. 
(c) Coarse-grained student ensemble absorbs a whole student without further considering fine-grained fragments in the model.
}
\label{Figure 1}
\end{figure}

To solve this problem, Distantly-Supervised Named Entity Recognition (DS-NER) has attracted increasing attention. It automatically annotates training data based on external knowledge such as easily-obtained dictionaries and knowledge bases, which effectively relieves the annotation difficulty. 
Unfortunately, such a distant labeling procedure naturally introduces incomplete and inaccurate labels. As depicted in Figure \ref{Figure 1} (a), ``White House" is unlabeled because the distant supervision source has limited coverage of the entity mentions. Meanwhile, ``Washington" is inaccurately labeled as this entity belongs to location types in the distant supervision source. 
Due to the existence of such noise in the distantly labeled data, straightforward application of supervised learning will yield deteriorated performance as deep neural models have a strong capacity of fitting the given noisy data. Thus, the robustness and generalization of learned DS-NER models are restricted. 

To address the above challenges, several DS-NER models have been proposed. \citet{shang2018learning} obtained high-quality phrases and designed TieOrBreak architecture to model those phrases that may be potential entities. \citet{peng2019distantly} adopt PU learning to perform classification using only limited labeled positive data and unlabeled data. However, these works mainly focus on designing network architectures that can cope with the incomplete annotations to partially alleviate the impact of the noisy annotations. 
Recently, the self-training teacher-student framework is applied to DS-NER tasks \cite{liang2020bond,zhang-etal-2021} to reduce the negative effect of both incomplete and inaccurate labels. 
This self-looping framework first selects high-confidence annotations from noisy labels to train the student network, and then updates a new teacher by the trained student. In this way, the training labels are gradually refined and  model generalization can be improved. 

However, the above self-training methods have the following shortcomings: (1) inadequate student learning. As shown in Figure \ref{Figure 1} (b), previous methods only focus on the consistent prediction from two teachers \cite{zhang-etal-2021} or simply consider the high-confidence part from a single teacher \cite{liang2020bond}. In this way, these models tend to learn uncomplicated mentions, and the entity recall rate will decrease. (2) coarse-grained teacher updating. In Figure \ref{Figure 1} (c), previous works absorb a whole student by exponential moving average (EMA) \cite{zhang-etal-2021} or directly copy the student as a new teacher \cite{liang2020bond} when updating the teacher. Such coarse-grained ensemble methods treat each model fragment equally while the noise sensitivity is diverse among different model fragments. 

In this paper,  we try to reconcile the above shortcomings with our newly proposed \textbf{A}daptive \textbf{T}eacher {L}earning and Fine-grained \textbf{S}tudent \textbf{EN}semble (ATSEN) for DS-NER. 
Specifically, we first apply two teacher networks to provide multi-view predictions on training samples. Then we propose an adaptive teacher learning which supervises agreement predictions by cross-entropy loss and accommodates disagreement parts with adaptive distillation. In this way, the student can be trained with more comprehensive knowledge. 
Subsequently, we update the new teacher with a fine-grained student ensemble, which updates a fragment of the teacher model with a temporal moving average of the corresponding fragment of the student. Therefore, the teacher model achieves more robustness for noise. 
With both adaptive learning and fine-grained ensemble, ATSEN is more effective than previous methods.  
We evaluate ATSEN on four DS-NER datasets. Experimental results demonstrate that our method significantly outperforms previous approaches.

To sum up, the main contributions of this paper are: 
\begin{itemize}
    \item To our best knowledge, this paper presents the first attempt to explore both agreement and conflicts among multiple teachers for the DS-NER by adaptive teacher learning, promoting comprehensive student learning.
    \item To further enhance the consistent prediction of model fragments, we devise a novel fine-grained student ensemble that stitches different fragments of previous student models into a unity. 
    In this way, the updated teacher achieves a more robust generalization ability.
    \item On four benchmark DS-NER datasets (Conll03, OntoNotes 5.0, WebPage, and Twitter), our ATSEN outperforms existing approaches by significant margins.

\end{itemize}

\begin{figure*}[tp]
\centering
\includegraphics[width=16cm]{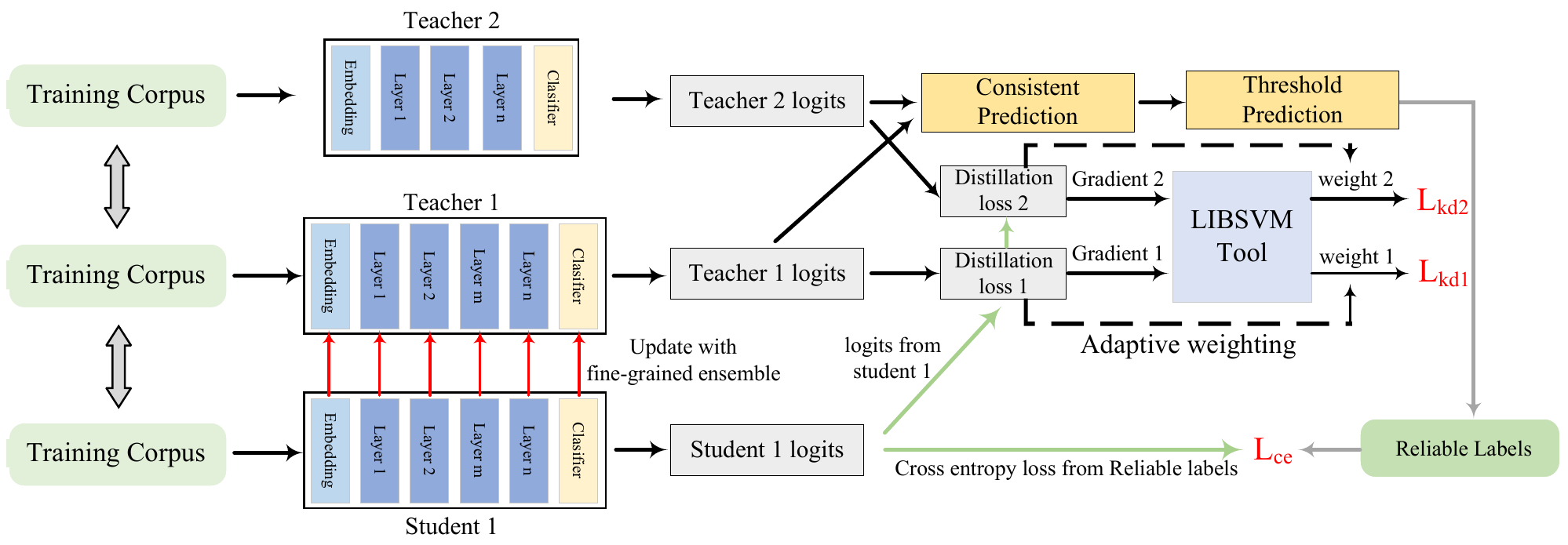}
\caption{Overview of our proposed ATSEN. Only the updating process of student 1 and teacher 1 is shown and the renewing process of student 2 and teacher 2 is similar. 
Specifically, the training corpus is first fed to two teachers and one student to obtain corresponding logits. Then the reliable labels are selected to supervise the student with cross-entropy loss. Meanwhile, adaptive distillation is adopted to further consider the disagreement tokens between teachers. Subsequently, a fine-grained ensemble is applied to the trained students to obtain a new teacher model. 
}
\label{Figure 2}
\end{figure*}

\section{Related Work}
Traditionally, many works have been proposed for supervised named entity recognition. For instance, \citet{huang2015bidirectional} utilized the BiLSTM as an encoder to learn the contextual representation and then exploited Conditional Random Field (CRF) as a decoder to label the tokens. More recently, deep learning methods \cite{xiao2020fine,qu2019adversarial} are introduced to different NLP fields, and strong pre-trained language models such as ELMo \cite{peters-etal-2018-deep} and BERT \cite{devlin2018bert} are incorporated to further enhance the performance of NER. However, most of these works rely on high-quality labels, which are expensive. Meanwhile, the reliance on labeled data also limits their applications in open situations. 

\noindent\textbf{DS-NER}
To address the labeled data scarcity problem, distantly-supervised named entity recognition methods are proposed. 
AutoNER \cite{shang2018learning} proposed a sequence labeling framework TieOrBreak and modify the standard CRF for adapting to the scenario of label noise. \citet{cao-etal-2019-low} promoted the quality of data by exploiting
labels in Wikipedia. AdaPU \cite{peng2019distantly} employed Positive-Unlabeled Learning to obtain unbiased estimation of the loss value. Conf-MPU \cite{zhou-etal-2022-distantly} further formulated the DS-NER problem via Multi-class Positive and Unlabeled (MPU) learning. 
BOND \cite{liang2020bond} adopted a teacher-student network to drop distant labels and use pseudo labels to gradually improve the model generalization ability. Similar to BOND, SCDL \cite{zhang-etal-2021} co-trained two teacher-student networks to form inner and outer loops for coping with label noise. In this paper, we propose a novel self-training framework to adaptively learn from multiple teachers and achieve a fine-grained student ensemble. In this way, our method achieves a more robust ability for noise in the DS-NER task.

\noindent\textbf{Teacher-Student Framework} The teacher-student framework is a popular architecture in many semi-supervised \cite{huo2021atso} and self-supervised \cite{abbasi2020compress} learning tasks, as well as knowledge distillation \cite{hinton2015distilling}. Recently, teacher-student framework attracts increasing attention in both computer vision \cite{he2020momentum,grill2020bootstrap} and natural language processing \cite{liang2020bond,zhang-etal-2021}. The teacher selects reliable annotations with devised strategies for student training and then the new teacher is updated based on the trained student. The optimization goal is to ensure the prediction consistency between the student and the teacher. In particular, there are several variants of teacher-student networks proposed for DS-NER. BOND devised a self-training teacher-student strategy that copies the student as a new teacher. With this self-training loop, the training pseudo labels are gradually refined. To improve the quality of pseudo labels and remove noise, SCDL designs two teachers and reaches an agreement between them to generate pseudo labels. Meanwhile, SCDL uses exponential moving average (EMA) to update the teacher based on the re-trained student. Following the self-training framework, we further improve the training process of both the teacher and student network to alleviate the noise problem. 

\section{Preliminaries}
Here we briefly describe the task definition of DS-NER. Formally, given the training data $D$, where each sentence is denoted as $(X^i, Y^i)$. $X^i$ is a token list that represents each word, and $Y^i$ is the corresponding tag list in the form of BIO schema. For DS-NER, we do not have access to human-annotated true labels, but only distant labels by matching unlabeled sentences with external dictionaries or knowledge bases (KBs). Thus, $Y^i$ may not be the underlying correct one. To generate distant labels, 
in this work, we follow the previous work \cite{liang2020bond}. 
The biggest challenge in DS-NER is how to reduce the label noise in the training samples and train a robust NER model as there is much ambiguity and limited coverage over entity types.

\section{Method}
In this work, considering the memory capacity and model efficiency, we train two sets of teacher-student networks instead of more pairs while our method can easily extend to more pairs.
The main procedure is shown in Figure 2.  

\subsection{Overall Framework}
The training procedure can be divided into three stages: 

\noindent (1) \textbf{Pretraining with initial noisy labels.} In this stage, we train two NER models ($\theta_1$, $\theta_2$) using the distant labels. These two models have different architectures in this work. Then, we duplicate these two models for the initialization of two sets of teacher networks, namely $\theta_{t1}=\theta_1$ and $\theta_{t2}=\theta_2$. 

The training target of $\theta_1$ and $\theta_2$ is:
\begin{equation}{
    L(\theta) =  -\frac{1}{MN}\sum^{M}_{i=1}\sum^N_{j=1}\overset{*}{y}_j^i log(p(y_j^i|X^i;\theta)) 
    }
\end{equation}

\noindent where $M$ is the number of sentences in the training corpus and $N$ is the token number in each sentence. ${\overset{*}{y}}_j^i$ means the distant label of $j$-th token of the $i$-th sentence.

\noindent (2) \textbf{Training student with adaptive teacher learning.} In this phrase, we select reliable labels by predictions of teachers from the first stage and supervise the students with cross-entropy loss. Meanwhile, considering the potential conflicts or competitions that exist among teachers, we investigate the diversity of teachers in the gradient space and recast the knowledge distillation from two teachers as a multi-objective optimization problem so that we can determine a better optimization direction for the training of student. To this end, an adaptive knowledge distillation loss is also adopted in this stage.

\noindent \textbf{Reliable Labels Selection.} Without any prior knowledge about which tokens are mislabeled or unlabeled, it is challenging to automatically detect them. Here we adopt two strategies to select reliable labels. (i) \textbf{Consistent Prediction.} The first token selection strategy is based on the pseudo labels prediction consistency between two teachers. 
\begin{equation}{
    (X^i, Y^i)_\text{CP} = \{(x_j, {y_j})| {y_j} = (y_{j,t1} == y_{j,t2})\}
    }
\end{equation}

\noindent where $y_{j,t1}, y_{j,t2}$ are predicted one-hot pseudo labels on training corpus for two teachers. If two teacher models predict the same labels on specific tokens, then the labels of these tokens are set to corresponding labels. Meanwhile, if two teacher models have different predictions, the labels of tokens will be set to the ``O" label.   
(ii) \textbf{Threshold Prediction}. We propose a simple threshold-based strategy to further filter reliable labels as  
the tokens with high confidence are more likely to be reliable. 
For teacher $t_1$, 
\begin{equation}
{
    (X^i, Y^i)_\text{TP} = \{ (x_j, {y_j}) | \text{max}({p_{j,t1}) > \sigma_1 \} }} 
\end{equation}

\noindent where $\sigma_1$ is the confidence threshold, ${p_{j,t1}}$ is the label distribution of the $j$-th token predicted by the teacher $t_1$. Thus, the tokens with label confidence lower than $\sigma_1$ will also be set to ``O" labels.  
After these two steps, we can obtain reliable labels $\overline{Y}$. 
With these reliable labels, we can supervise the student models with the cross-entropy loss as follows:

\begin{equation}{
    L_{ce}(\theta) =  -\frac{1}{MN}\sum^{M}_{i=1}\sum^N_{j=1}\overline{y}_j^i log(p(y_j^i|X_i;\theta)) 
    }
\end{equation}

\subsubsection{Adaptive Distillation}
The above selection procedure only considers consistent parts between two teachers while the conflicts among teachers are not squared up. 
To handle the inner conflicts, we formulate ensemble knowledge from teachers as a multi-objective optimization (MOO) problem \cite{sener2018multi} and use multiple gradient descent algorithms (MGDA) to probe a Pareto optimal solution that accommodates all teachers as much as possible. 

Specifically, we first formally introduce the standard knowledge distillation loss which encourages the logits of the student network to mimic the teacher network:
\begin{equation}
\begin{split}
    L_{kd}^t ({\theta}) = H(p^s, p^t) = H(\sigma(a^s; T), \sigma(a^t; T)) = \\ 
    -\sum_{k=1}^{K} p^t \text{log} p^s[k] = - \left \langle p^t, \text{log} p^s \right \rangle
\end{split}  
\end{equation}
where $\sigma$ is softmax operation, $a^s$ and $a^t$ are the logits of student and teacher networks, $T$ is the temperature to soften the logits. $K$ is the number of classification types. $H(\cdot, \cdot)$ is the cross-entropy loss to measure the discrepancy of softened probabilistic output between the student and teacher. 
In this work, we have two teachers, thus the naive solution for distilling from two teachers is:

\begin{equation}
\begin{split}
    L_{kd}({\theta}) = L_{kd}^{t1} ({\theta}) + L_{kd}^{t2} ({\theta}) = H(p^s, p^{t1}) + H(p^s, p^{t2})
\end{split}  
\end{equation}

However, conservatively accepting the directions from all teachers, \textit{i.e.}, accumulating the separate distillation loss from each teacher, is not a good option, since the diversity of teachers could be significant and there might be some weak or noisy teachers mingled in the ensemble. 
When distilling knowledge from multiple teachers, we need to incorporate the disagreement into the determination of the descent direction. 
Recently, a novel method is proposed to find one single Pareto optimal solution with a good trade-off among conflicting optimization targets. 
Following \cite{sener2018multi,lin2019pareto}, we can reformulate the Pareto solution of learning from two teachers as a linear scalarization of tasks with adaptive weight assignment as follows:

\begin{equation}
L(\theta) = \alpha_1 L_{kd}^{t1} + \alpha_2 L_{kd}^{t2}
\end{equation}

\noindent where we adaptively assign the weights $\alpha_m$ by solving the following problem in each iteration:
\begin{equation}
\begin{split}
\text{min} \  \frac{1}{2} || \sum_{m=1}^M \alpha_m \nabla_\theta L_{kd}^m (\theta^\tau)|| ^2 , s.t. \\
\sum_{m=1}^M \alpha_m = 1, \ 0 \leq \alpha_m \leq C, \forall m \in [1:M]
\end{split}
\end{equation}
where $C >0$ is the regularization parameter, and $M$ is the number of teachers. $L_{kd}^m (\theta^\tau)$ is the knowledge distillation loss at Eq. 5 corresponding to the student and $m$-th teacher. $\theta^\tau$ is the parameter of the student network at iteration $\tau$.
Considering that calculating the gradient over parameters $\theta^\tau$ can be fairly time-consuming. Following \cite{sener2018multi}, we turn to its upper bound:

\begin{equation}
{
\text{min} \ \frac{1}{2} ||  \sum_{m=1}^M \alpha_m \nabla_Z L_m (\theta^\tau) ||^2, s.t. 
}
\end{equation}

\noindent where $\sum_{m=1}^M \alpha_m = 1, \ 0 \leq \alpha_m \leq C, \forall m \in [1:M]$, $Z$ is the feature over the corresponding teacher. In this way, Eq. 9 is a typical One-class SVM problem and can be solved by LIBSVM \cite{chang2011libsvm}. More intuitively, as shown in Figure 2, we first compute the standard distillation loss according to the student logit and each teacher logit. Through the back-propagation algorithm, we can obtain the gradients corresponding to each teacher for the student model. Subsequently, we solve the loss weights through the gradients with the LIBSVM tool.   

Finally, the total training loss for the student model in the second stage is: 

\begin{equation}
{
L(\theta) = L_{ce}(\theta) + \alpha L_{kd}^{t1} ({\theta}) + (1-\alpha) L_{kd}^{t2} ({\theta})
}
\end{equation}

\noindent (3) \textbf{Updating teacher with fine-grained student ensemble.} 
After training the students, we devise a fine-grained student ensemble to update the parameters of the teachers. Before describing the concrete fine-grained ensemble, we first introduce a preliminary version, named \textbf{segment ensemble (SE)}. 
During each iteration, the segment ensemble picks up some units of the student model to replace the corresponding units of the teacher model, leaving the remaining parts of the teacher unchanged. Formally, at iteration $\tau$, 

\begin{equation}
{
{\theta}^t({\tau}) = \{ |P_i < {\sigma_2} | \theta_i^t({\tau}-1) + (1-|P_i < {\sigma_2} |)\theta_i^s({\tau}) \}
}
\end{equation}
where $P_i$ is random probability distribution in [0,1] for the i-th unit of the teacher which is independent of each other. If $P_i < \sigma_2$, then $|P_i < \sigma_2 | = 1$, the i-th unit parameter of teacher is to be preserved. In our paper, each unit corresponds to one network layer of the student network. 
The motivation of our segment ensemble is from Dropout \cite{srivastava2014dropout} while Dropout works when training a network. 

Subsequently, the segment ensemble can further integrate with EMA to incorporate temporal property. Here we first review the traditional EMA strategy:

\begin{equation}
{
\theta^t(\tau) = \{  m \theta ^ t (\tau-1) + (1 - m) \theta ^s (\tau)\}
}
\end{equation}
where $m$ denotes the smoothing coefficient. As shown in this equation, EMA treats the model as a whole. We can integrate these two ensemble methods as fine-grained ensemble:

\begin{equation}
\begin{aligned}
\theta^\tau(\tau) &= \{|P_i < \sigma_2 | \theta^t(\tau-1) + (1-|P_i < \sigma_2 |)m\theta_i^t(\tau-1) \\ 
&+ (1-|P_i < \sigma_2 |)(1-m)\theta_i^s(\tau)  \} 
\end{aligned}  
\end{equation}
when $m=0$, it becomes segment ensemble. Similarly, it degenerates to EMA when $ \sigma_2 = 0$. In this manner, the fine-grained ensemble not only possesses the temporal property of traditional EMA, but also enhances the robustness of each segment to noise.  
As a result, the teacher tends to generate more reliable pseudo labels, which can be used as new supervision signals in the next round self-training.

To sum up, the first stage is executed once for a moderate initialization with distant labels. The second and third phases will be conducted alternately in a loop for better student and teacher models. 
Finally, only the best model $\theta \in \{\theta_{t1}, \theta_{t2}, \theta_{s1}, \theta_{s2}\}$ will be used for prediction.

The details of our model are presented in Algorithm 1.

\begin{algorithm}[tb]
\small
\setstretch{1.15}
\captionsetup{font={small}}
\caption{ATSEN training.}
\label{alg:algorithm}
\textbf{Input}: Training corpus  $\mathcal{D}=\{(X_i, Y_i)\}_{i=1}^M$ with noisy labels \\
\textbf{Parameter}: Two network parameters $\theta_{t_1}$, $\theta_{s_1}$, $\theta_{t_2}$, and $\theta_{s_2}$ \\
\textbf{Output}: The best model
\begin{algorithmic}[1] 
\STATE Pre-training two models $\theta_1$, $\theta_2$ with $\mathcal{D}$. \hfill $\triangleright${\it Pre-Training}.\\
\STATE $\theta_{t_1} \gets \theta_1$, $\theta_{t_2} \gets \theta_2$, $step \gets 0$.\\
\STATE Initialize noisy labels: $Y_{\uppercase\expandafter{\romannumeral1}} \gets Y, Y_{\uppercase\expandafter{\romannumeral2}} \gets Y$.\\
\WHILE{ \emph{not reach max training epochs} }
    \STATE Get a batch $(X^{(b)}, Y^{(b)}_{\uppercase\expandafter{\romannumeral1}}, Y^{(b)}_{\uppercase\expandafter{\romannumeral2}})$ from $\mathcal{D}$, \\
    $step \gets step+1$. \hfill $\triangleright${\it Self-Training}.\\
    \STATE Get pseudo-labels via the teacher $\theta_{t_1}$, $\theta_{t_2}$: \\ $ \tilde{Y}^{(b)}_{\uppercase\expandafter{\romannumeral1}} \gets f(X^{(b)}; \theta_{t_1}) $, \\ $ \tilde{Y}^{(b)}_{\uppercase\expandafter{\romannumeral2}} \gets f(X^{(b)}; \theta_{t_2})$. \\
    \STATE Get reliable tokens by Eq. 2 and Eq. 3: \\ $\mathcal{T}^{(b)}_{\uppercase\expandafter{\romannumeral1}} \gets$ TokenSelection$(Y^{(b)}_{\uppercase\expandafter{\romannumeral1}},  \tilde{Y}^{(b)}_{\uppercase\expandafter{\romannumeral1}})$, \\ $\mathcal{T}^{(b)}_{\uppercase\expandafter{\romannumeral2}} \gets$ TokenSelection$(Y^{(b)}_{\uppercase\expandafter{\romannumeral2}},  \tilde{Y}^{(b)}_{\uppercase\expandafter{\romannumeral2}})$. \\
    \STATE Update the student $\theta_{s_1}$ and $\theta_{s_2}$ by Eq. 10. \\
    \STATE Update the teacher $\theta_{t_1}$ and $\theta_{t_2}$ by Eq. 13. \\
        \STATE Update noisy labels mutually: \\
        $Y_{\uppercase\expandafter{\romannumeral1}}=\{Y_i \gets f(X_i; \theta_{t_2})\}_{i=1}^M$, \\
        $Y_{\uppercase\expandafter{\romannumeral2}}=\{Y_i \gets f(X_i; \theta_{t_1})\}_{i=1}^M$.

\ENDWHILE
\STATE Evaluate models $\theta_{t_1}$, $\theta_{s_1}$, $\theta_{t_2}$, $\theta_{s_2}$ on {\it Dev} set. \\
\STATE \textbf{return} The best model $\theta\in\{ \theta_{t_1}, \theta_{s_1}, \theta_{t_2}, \theta_{s_2}\}$
\end{algorithmic}
\end{algorithm}

\section{Experiments}

\begin{table}[thp]
\renewcommand\arraystretch{1.1}
\centering
\scriptsize
\begin{tabular}{cccccc}
\toprule
\multicolumn{2}{c}{\textbf{Dataset}} & Train  & Dev  & Test  & Types \\ \midrule
\multirow{2}{*}{\textbf{CoNLL03}} & \textbf{Sentence}     & 14041      & 3250   & 3453  & \multirow{2}{*}{4}  \\
& \textbf{Token}            & 203621  & 51362  & 46435 \\ \hline
\multirow{2}{*}{\textbf{OntoNotes5.0}} & \textbf{Sentence}     & 115812     & 15680  & 12217  & \multirow{2}{*}{18}  \\
& \textbf{Token}            & 2200865 & 304701 & 230118 \\ \hline
\multirow{2}{*}{\textbf{Webpage}} & \textbf{Sentence}     & 385     & 99  & 135  & \multirow{2}{*}{4}  \\
& \textbf{Token}            & 5293 & 1121 & 1131 \\ \hline
\multirow{2}{*}{\textbf{Twitter}} & \textbf{Sentence}     & 2393     & 999  & 3844  & \multirow{2}{*}{10}  \\
& \textbf{Token}            & 44076 & 15262 & 58064 \\ \bottomrule
\end{tabular}
\caption{The statistics of four DS-NER datasets.}
\label{appendix Table 1}
\end{table}

\begin{table*}[t!]

    \centering
    \resizebox{16.5cm}{2.95cm}{
    \begin{tabular}{ccccccccccccc}
    \hline 
    \multirow{2}*{Method} & \multicolumn{3}{c}{CoNLL03} & \multicolumn{3}{c}{OntoNotes 5.0} & \multicolumn{3}{c}{Webpage} & \multicolumn{3}{c}{Twitter} \\ \cline{2-4} \cline{5-7} \cline{8-10} \cline{11-13}
    ~ & P & R & F1 & P & R & F1 & P & R & F1 & P & R & F1 \\ \hline

{BiLSTM-CRF}\textsuperscript{$\clubsuit$}  & 91.35 & 91.06 & 91.21 & 85.99 & 86.36 & 86.17 & 50.07 & 54.76 & 52.34 & 60.01 & 46.16 & 52.18 \\
{RoBERTa}\textsuperscript{$\clubsuit$*}  & 90.61 & 91.72  & 91.22 & 84.59 &87.88 & 86.20  & 66.29 &79.73 & 72.39 & 57.32 & 51.85 & 54.45 \\ 
\hline

{KB-Matching}  & 81.13 & 63.75 & 71.40 & 63.86 & 55.71 & 59.51 & 62.59 &45.14 & 52.45 & 40.34 & 32.22 & 35.83 \\
{Diatant BiLSTM-CRF} & 75.50 & 49.10 & 59.50 & \textbf{68.44} & 64.50 & 66.41 & 58.05 & 34.59 & 43.34 & 46.91 & 14.18 & 21.77 \\
{Distant DistilRoBERTa} & 77.87 & 69.91 & 73.68 & 66.83 & 68.81 & 67.80 & 56.05  & 59.46  & 57.70 & 45.72 & 43.85 & 44.77 \\
{Distant RoBERTa} & 82.29       & 70.47       & 75.93       & 66.99       & 69.51       & 68.23       & 59.24       & 62.84       & 60.98        & 50.97       & 42.66       & 46.45      \\

{AutoNER} & 75.21       & 60.40        & 67.00          & 64.63       & 69.95           & 67.18       & 48.82       & 54.23       & 51.39    & 43.26       & 18.69       & 26.10       \\
{LRNT}    & 79.91       & 61.87       & 69.74       & 67.36       & 68.02       & 67.69       & 46.70        & 48.83       & 47.74       & 46.94       & 15.98       & 23.84      \\
{Co-teaching+}          & 86.04       & 68.74       & 76.42       & 66.63       & 69.32       & 67.95       & 61.65       & 55.41       & 58.36             & 51.67       & 42.66       & 46.73      \\
{NegSampling}    & 80.17       & 77.72             & 78.93       & 64.59       & \textbf{72.39}       & 68.26        & {70.16}       & 58.78       & 63.97           & 50.25       & 44.95             & 47.45 \\
{BOND}   & 82.05       & {80.92}             & 81.48       & 67.14       & 69.61       & 68.35       & 67.37       & 64.19       & 65.74            & 53.16       & 43.76             & 48.01       \\ 
{SCDL}             & \textbf{87.96}             & 79.82       & {83.69}             & 67.49       & 69.77       & {68.61}             & 68.71             &  {68.24}             & {68.47}                 & {59.87}             & 44.57       &  {51.09}            \\ \hline
\textbf{ATSEN}             & {86.14}             &  \textbf{85.05}       & \textbf{85.59}             & 66.97       & {71.05}       & \textbf{68.95}             &  \textbf{71.08}             &  \textbf{70.03}             &  \textbf{70.55}                 & \textbf{62.32}             & \textbf{45.30}       & \textbf{52.46}     \\ \hline

\end{tabular}}
\caption{Main results on four benchmark datasets measured by precision (P), recall (R) and F1 scores. Baselines are reported by \cite{zhang-etal-2021}. $\clubsuit$ marks the model trained on the fully clean dataset. * denotes models implemented by us.}
    \label{Table 2}
\end{table*}

\subsection{Datasets}
To verify the effectiveness of our proposed ATSEN, we conduct experiments on four DS-NER datasets. Here we give a short description of them as follows:

\noindent\textbf{CoNLL03} \cite{sang2003introduction} consists of 1393 English news articles and is annotated with four entity types: person, location, organization, and miscellaneous.

\noindent\textbf{OntoNotes 5.0} \cite{weischedel2013ontonotes} contains documents from multiple domains, including broadcast conversation, P2.5 data, and Web data. It consists of 18 entity types.

\noindent\textbf{Webpage} \cite{ratinov2009design} comprises of personal, academic, and computer science conference webpages. It consists of
20 webpages that cover 783 entities. 

\noindent\textbf{Twitter} \cite{godin2015multimedia} is from the WNUT 2016 NER shared task. It consists of 10 entity types.

\noindent The detailed statistics of each dataset are listed in Table 1. 

\subsection{Compared Methods}
We compare our ATSEN with a wide range of state-of-the-art DS-NER methods and supervised methods. Fully supervised methods use the ground truth annotation for model training. DS-NER methods use the distantly-labeled training set provided in \cite{liang2020bond}.

\noindent\textbf{Fully-supervised Methods.} We include two supervised NER methods for comparison. (1) RoBERTa \cite{liu2019roberta} adopts RoBERTa model as backbone and a top linear layer for token-level classification. (2) BiLSTM-CRF \cite{ma2016end} uses bi-directional LSTM with character-level CNN to produce token embeddings, which are then fed into a CRF layer to predict token labels.

\noindent\textbf{Distantly-supervised Methods.} (1) KB-Matching reports the distant supervision quality. (2) Distant BiLSTM-CRF, Distant DistilRoBERTa, and Distant RoBERTa fine-tune the corresponding models on distantly-labeled data as if they are ground truth with the standard supervised learning. (3) AutoNER \cite{shang2018learning} trains the model by assigning
ambiguous tokens with all possible labels and then maximizing the overall likelihood using a fuzzy CRF model. LRNT \cite{cao-etal-2019-low} applies partial-CRFs on high-quality data with non-entity sampling. Co-teaching+ \cite{yu2019does} is a classic de-nosing method in computer vision. NegSampling \cite{li2020empirical} only handles incomplete annotations by negative sampling. BOND and SCDL both adopt self-training strategies that are straightforward competitors to ATSEN.

\subsection{Implementation Details}
The architecture of the teachers is the backbone language model and a top classification layer for token-level classification. Specifically, we adopt RoBERTa and DistilRoBERTa as backbone for teacher 1 and teacher 2.  
The corresponding student has the same architecture as their teacher. The max training epoch is 50 for all datasets. The training batch size is 16 for CoNLL03, Webpage, and Twitter and 32 for OntoNotes 5.0. The learning rate is set to 1e-5 for CoNLL03 and Webpage, and 2e-5 for OntoNotes 5.0 and Twitter. 
For the pretraining stage with noisy labels, we separately train 1, 2, 12, and 6 epochs for CoNLL03, OntoNotes 5.0, Webpage, and Twitter datasets. For adaptive teacher learning, the confidence threshold $\sigma_1$ is 0.9 for all datasets. 
In the fine-grained student ensemble, $m$ are 0.995, 0.995, 0.99, 0.995 and $\sigma_2$ is set to 0.8, 0.995, 0.8, and 0.75 for dataset CoNLL03, OntoNotes 5.0, Webpage, and Twitter, respectively.

\subsection{Main Results}
Table 2 presents the performance of all methods measured by precision, recall, and F1 scores. The results are summarized as follows: On all four datasets, ATSEN achieves the best performance among all distantly-supervised methods. Specifically, the distant DistilRoBERTa and RoBERTa only slightly improve the distant labeling performance compared to the naive KB-Matching, showing that directly applying supervised learning to distantly-labeled data will lead to poor model generalization. 
In addition, ATSEN performs much better than
previous studies which consider the noisy labels in
NER, including AutoNER, LRNT, Co-teaching+, and NegSampling.
When compared to strong self-training methods BOND and SCDL, our ATSEN achieves new state-of-the-art performance, demonstrating the superiority of our proposed adaptive teacher learning and fine-grained student ensemble when trained on distantly-labeled data.
Concretely, on CoNLL03, ATSEN achieves 1.90 absolute F1 improvements over the strong method SCDL. On the biggest and most difficult dataset OntoNotes V5.0, we obtain a decent improvement compared to the SOTA approach SCDL by 0.34 F1 score. In addition, we get 2.08 and 1.37 F1 scores improvement on Webpage and Twitter respectively.

\begin{table}[t]
\centering
\begin{tabular}{l*{3}{c}}
\toprule
\textbf{Ablations} & \textbf{Precision} & \textbf{Recall} & \textbf{F1} \\
\midrule
\textbf{ATSEN} & 86.14 & 85.05 & 85.59 \\
\qquad \qquad\textbf{w/o CP} & 83.48 & 81.67 & 82.56 \\
\qquad \qquad\textbf{w/o TP} & 84.66 & 85.83 & 83.54 \\
\qquad \qquad\textbf{w/o CE} & 86.05 & 84.37 & 85.20 \\
\qquad \qquad\textbf{w/o AD} & 87.96 & 82.14 & 84.95 \\
\qquad \qquad\textbf{w/o FE} & 83.57 & 84.66 & 84.11 \\
\bottomrule
\end{tabular}
\caption{
Ablation study on CoNLL03 dataset. We compare ATSEN with ablations by removing specific components.
}
\label{tab:ablation}
\end{table}

\subsection{Ablation study}
To further validate the effectiveness of each component in our ATSEN, we compare ATSEN with the following ablations by removing specific components: (1) remove the consistent prediction (w/o CP) in Eq.2. (2) remove the threshold prediction (w/o TP) in Eq.3. 
(3) do not perform cross-entropy loss (w/o CE).
(4) do not perform adaptive distillation (w/o AD), namely, only cross-entropy loss is adopted in Eq.10.
(5) do not perform fine-grained ensemble (w/o FE), namely, directly copy the trained student as a new teacher. 

As shown in Table 3, it can be observed that w/o CP and w/o TP lead to a significant performance drop, indicating these strategies are important for cross-entropy learning. Meanwhile, the result of w/o CE do not cause huge performance as adaptive distillation also considers the agreement part between teachers. The results from w/o CP, w/o TP, and w/o CE also imply that the cross-entropy loss from ambiguous labels may damage the performance.   

In addition, the result of w/o AD decreases the recall largely compared to ATSEN. It shows that considering knowledge from the disagreement part of two teachers can effectively help comprehensive student learning.  
Finally, w/o FE significantly reduces performance, showing that our fine-grained ensemble indeed benefits the model's generalization ability.

\begin{table}[t]
\centering
\begin{tabular}{lc}
\toprule
\text{Ablations} & \text{F1 Score} \\
\midrule
\text{Baseline (adaptive distillation)} & 85.59  \\
\text{average distillation} & 85.19  \\
\text{manually weighted distillation} & 85.22 \\
\text{dynamically weighted distillation} & 85.26 \\
\text{disagreement distillation} & 84.86 \\
\bottomrule
\end{tabular}
\caption{
F1 scores of different variants on CoNLL03 dataset. 
}
\label{tab:ensemble}
\end{table}

\subsection{Study of Adaptive Distillation}
In this section, we study the effectiveness of adaptive distillation for the student training process. Here we implement several ablations as shown in Table 4. 
The baseline is the adaptive teacher learning used in Eq.10, where 
the weight $\alpha$ is computed by LIBSVM from the gradients of two teachers corresponding to the student. 
We devise four variants. The first variant is averaging the distillation loss to substitute the adaptive distillation. It decreases by about 0.4 F1 scores. We also devise the second variant by manually setting the weights. Here we set 0.7 and 0.3 for distillation loss from teacher 1 (Roberta) and teacher 2 (DistilRoBERTa). This variant still performs worse than adaptive distillation. We have also tried other weight combinations such as 0.8 and 0.2 but achieved even worse results. Furthermore, we directly learn a dynamic weight $\alpha$ and achieve similar results with the manual setting. 
Finally, we consider a variant that only considers the disagreement part between two teachers during distillation, thus the training tokens may not be continuous and complete. The result presents a performance drop, indicating that the gradient optimization for adaptive distillation should be conducted for the whole sentence.

\begin{figure}[tp]
  \centering
  {\includegraphics[width=0.99\columnwidth]{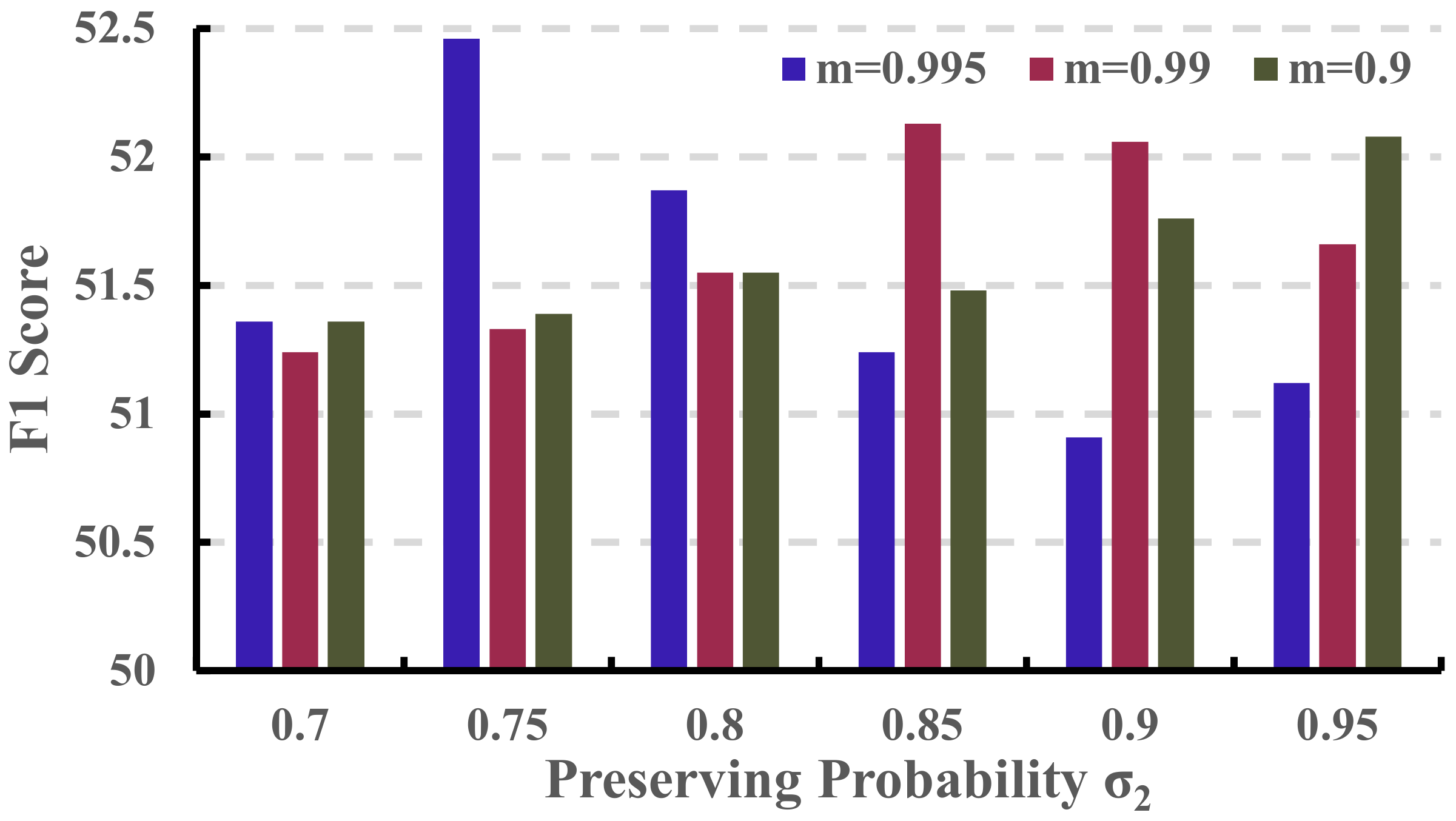}}
  \label{fig5}
  \caption{F1 score on Twitter dataset of different variants.}
\end{figure}

\subsection{Study of Fine-grained Ensemble}
We investigate the effectiveness of different student ensemble methods. For comprehensive evaluation, we experiment on a relatively smaller dataset Twitter instead of CoNLL03. As shown in Table 5: (1) remove all ensemble strategies (w/o all) and directly copy the student as a new teacher. (2) remove the segment ensemble (w/o SE), namely $\sigma_2=0$ in Eq.13. (3) remove the EMA (w/o EMA), namely $m=0$ in Eq.13. As shown in Table 5, w/o all lead to the most significant performance drop. Meanwhile, removing either SE or EMA cause decreased results, demonstrating these two kinds of ensemble method can complement each other. It is worth noting ATSEN achieves significantly better precision than variants, indicating fine-grained ensemble can effectively enhance consistent predictions by performing on model fragments.   
Furthermore, we investigate the parameter influence of fine-grained ensemble in Fig. 3. As shown in this figure, we can observe $m=0.995$ and $\sigma_2=0.75$ achieve the best performance. We also notice 
an interesting fact is that with the increase of $m$, the model achieves its best performance at a relatively smaller value of $\sigma_2$.

\subsection{Case Study}
We perform case study to understand the advantage of our proposed ATSEN with a concrete example in Table 6. We show the prediction result of BOND, SCDL, and ATSEN on a training sequence with label noise. 
BOND can slightly generalize to unseen mentions and relieve partial incomplete annotation. For example, BOND can locate the ``John McNamara" and ``New York" while distant labels only can match partial person names.  
SCDL is able to generalize better for more accurate entity detection because it has a co-training step. For instance, SCDL can further locate the entity ``Columbia Presby Hospital". However, it is still impacted by label noise. For comparision, for hard labels ``California Angels", 
our ATSEN is able to detect them with both adaptive teacher learning and fine-grained student ensemble, instead of relying purely on distant labels. 

\begin{table}[t]
\centering
\begin{tabular}{l*{3}{c}}
\toprule
\textbf{Ablations} & \textbf{Precision} & \textbf{Recall} & \textbf{F1} \\
\midrule
\textbf{ATSEN} & 62.32 & 45.30 & 52.46 \\
\qquad \qquad\textbf{w/o all} & 55.85 & 42.30 & 48.14 \\
\qquad \qquad\textbf{w/o SE} & 58.90 & 45.35 & 51.25 \\
\qquad \qquad\textbf{w/o EMA} & 59.67 & 46.65 & 52.36 \\
\bottomrule
\end{tabular}
\caption{
Ablation study on Twitter dataset. We compare our full method ATSEN with several ensemble strategy variants.
}
\label{tab:ablation}
\end{table}

\newcommand{\per}[1]{\textcolor{red}{[#1]$_{\text{PER}}$}}
\newcommand{\org}[1]{\textcolor{cyan}{[#1]$_{\text{ORG}}$}}
\newcommand{\loc}[1]{\textcolor{blue}{[#1]$_{\text{LOC}}$}}

\begin{table}[t]
\centering
\scalebox{0.85}{
\begin{tabular}{l}
\toprule
\textbf{Distant}: \per{Johnson} is the second manager to be hospitalized \\ after 
California \per{Angels} skipper 
 \per{John}  McNamara \\ was admitted to New \per{York} 's \per{Columbia} Presby Hospital. \\ 
\textbf{Golden}:  \per{Johnson} is the second manager to be hospitalized \\ after
\org{California Angels} skipper 
 \per{John  McNamara} \\ was admitted to \loc{New York} 's \org{Columbia Presby Hospital}. \\
\midrule
\textbf{BOND}:  \per{Johnson} is the second manager to be hospitalized \\ after
\loc{California} \per{Angels} skipper 
 \per{John McNamara} \\ was admitted to \loc{New York} 's \per{Columbia} Presby Hospital. \\
\textbf{SCDL}:  \per{Johnson} is the second manager to be hospitalized \\ after
\loc{California} \per{Angels} skipper 
 \per{John  McNamara} \\ was admitted to \loc{New York} 's \org{Columbia Presby Hospital}. \\
\textbf{ATSEN}:  \per{Johnson} is the second manager to be hospitalized \\ after
\org{California Angels} skipper 
 \per{John  McNamara} \\ was admitted to \loc{New York} 's \org{Columbia Presby Hospital}. \\
\bottomrule
\end{tabular}
}
\caption{
Case study. The sentence is from CoNLL03 dataset.
}
\label{tab:case}
\end{table}

\section{Conclusion}
In this paper, we present a novel self-training framework ATSEN for DS-NER. Specifically, ATSEN adopts adaptive teacher learning to train student networks, considering both consistent and inconsistent predictions between them. Furthermore, we devise a fine-grained student ensemble to update the teacher model. With it, each fragment of the teacher benefits from a temporal moving average of the corresponding fragment of the student.
The experiment results illustrate that ATSEN significantly outperforms SOTA methods.

\section{Acknowledgments}
Pan Zhou is funded by the National Natural Science Foundation of China (NSFC) with grant numbers 61972448.

\bibliography{aaai23}

\end{document}